\PassOptionsToPackage{table}{xcolor} 
\PassOptionsToPackage{dvipsnames}{xcolor} 
\documentclass[10pt,twocolumn,letterpaper]{article}

\usepackage[pagenumbers]{iccv}              
\usepackage{adjustbox}
\usepackage{fontawesome5} 
\usepackage{bbding}
\usepackage{multirow} 
\usepackage{algorithm}
\usepackage{algpseudocode}
\usepackage{lipsum}
\definecolor{lightblue}{rgb}{0.95, 0.95, 1}
\usepackage[table]{xcolor}
\usepackage{bm}
\usepackage{comment}

\usepackage{bm}

\definecolor{iccvblue}{rgb}{0.21,0.49,0.74}
\usepackage[pagebackref,breaklinks,colorlinks,allcolors=iccvblue]{hyperref}

\def\paperID{11502} 
\def\confName{ICCV}
\def\confYear{2025}

\title{PRISM: Reducing Spurious Implicit Biases in Vision-Language Models with LLM-Guided Embedding Projection}

\author{
Mahdiyar Molahasani\thanks{These authors contributed equally.} , Azadeh Motamedi\footnotemark[1] , Michael Greenspan, Il-Min Kim, Ali Etemad 
\\
Queen’s University, Canada\\
{\tt\small \{m.molahasani, 19am43, michael.greenspan, ilmin.kim, ali.etemad\}@queensu.ca}
}

\begin{document}
\maketitle

\begin{abstract}
We introduce Projection-based Reduction of Implicit Spurious bias in vision-language Models (PRISM), a new 
data-free and task-agnostic solution for bias mitigation in VLMs like CLIP. 
VLMs often inherit and amplify biases in their training data, leading to skewed predictions.
PRISM is designed to debias VLMs without relying on predefined bias categories or additional external data. It operates in two stages: first, an LLM is prompted with simple class
prompts to generate scene descriptions that contain spurious correlations. Next, PRISM uses our 
novel contrastive-style debiasing loss to learn a projection that maps the embeddings onto a latent space that minimizes spurious correlations while preserving the alignment between image and text embeddings. 
Extensive experiments demonstrate that PRISM outperforms current debiasing methods on the commonly used Waterbirds and CelebA datasets
We make our code public at: \url{https://github.com/MahdiyarMM/PRISM}.
\end{abstract}

\section{Introduction}
Vision-Language Models (VLMs) have emerged as a class of machine learning models capable of understanding and generating multi-modal content by jointly processing images and textual descriptions. Prominent VLMs, such as Contrastive Language-Image Pre-training (CLIP)
\cite{radford2021learning}, have demonstrated impressive performance in zero-shot classification and cross-modal retrieval. 
By aligning textual and visual representations in a shared embedding space, these models enable seamless interactions between vision and language tasks. Despite their unprecedented effectiveness, VLMs inherit and propagate systemic biases from their large-scale pretraining data, leading to inaccurate or skewed predictions. 

\begin{figure}[t!]
  \centering
      \begin{subfigure}{0.49\linewidth}
    \includegraphics[width=1.02\linewidth]{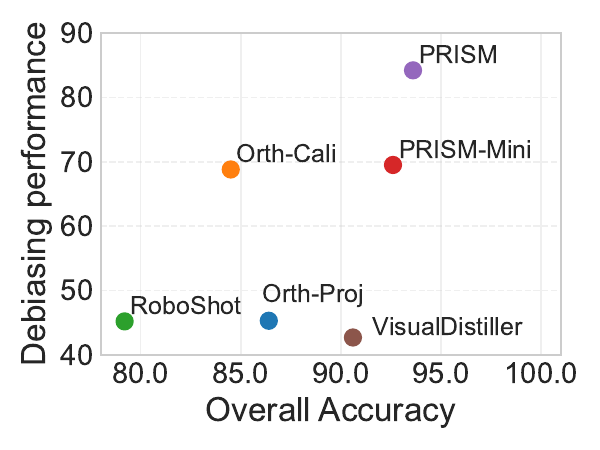}
    \caption{Waterbirds.}
    \label{fig:tradeoff-a}
  \end{subfigure}
  \hfill
  \begin{subfigure}{0.49\linewidth}
    \includegraphics[width=1.02\linewidth]{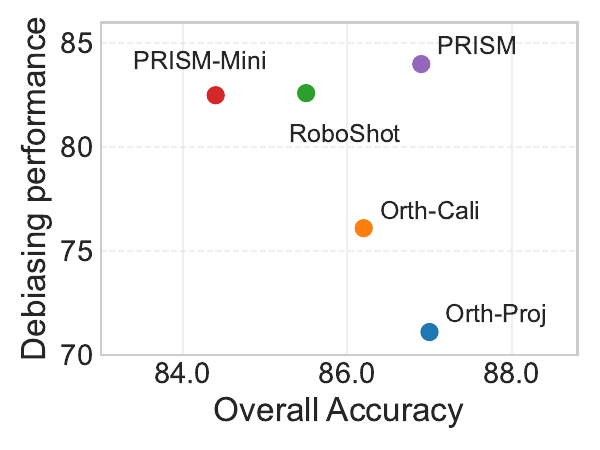}
    \caption{CelebA.}
    \label{fig:tradeoff-b}
  \end{subfigure} 
  \caption{Performance comparison among data-free CLIP debiasing methods in terms of debiasing performance (i.e., the accuracy of the worst group) vs. the overall accuracy on (a) Waterbirds and (b) CelebA datasets.
  Our proposed method, PRISM, achieves state-of-the-art debiasing performance (higher worst-group accuracy) without compromising overall accuracy.
  }
  \label{fig:short}
\end{figure}

Biases in VLMs manifest in various forms, including social \cite{zhou2022vlstereoset, meister2023gender, yang2020towards, wang2021gender} and contextual biases \cite{zeng2024understanding, wang2022revise}, often influenced by `spurious correlations' in training data. 
Spurious correlations can stem from associating non-target attributes, such as image backgrounds or contextual features, with labels in unintended ways \cite{chuang2023debiasing}. 
For instance, CLIP’s zero-shot classifiers frequently associate
`camel' with `desert' and `panda' with `bamboo', 
reflecting underlying biases in training data. These biases can have real-world implications 
reducing model robustness in diverse settings. Therefore, mitigating implicit bias in VLMs has become a critical research direction.

Recent research has investigated various strategies to \textit{identify}, \textit{quantify}, and \textit{reduce} spurious correlations in VLMs. 
These approaches can be classified into three categories: feature-space transformations \cite{dehdashtian2024fairerclip,wortsman2022robust}, prompt-based debiasing \cite{chuang2023debiasing, dai2024refining, adila2024zero}, and adapter design \cite{ kirichenko2022last}.
Methods with a primary focus on the feature space
\cite{dehdashtian2024fairerclip, wortsman2022robust} attempt to transform feature embeddings to eliminate spurious correlations. Prompt-based techniques \cite{chuang2023debiasing, dai2024refining, adila2024zero} aim to redefine textual inputs to mitigate bias but often struggle with generalization. Meanwhile, adapter-based approaches such as \cite{kirichenko2022last} address implicit bias in VLMs by modifying representations at various stages of the model, including but not limited to the final layer. 

Despite the effectiveness of these approaches in addressing spurious bias in VLMs, they fall short in many instances. 
These methods mostly require fixed predefined bias categories makes them difficult to scale to task-agnostic settings \cite{dehdashtian2024fairerclip,chuang2023debiasing, dai2024refining}.
Furthermore, these debiasing solutions either compromise the overall performance of the model by removing essential semantic content \cite{adila2024zero, chuang2023debiasing} or rely on external visual data for debiasing \cite{dehdashtian2024fairerclip, kirichenko2022last,wortsman2022robust}. This underscores the need for a new debiasing method without the reliance on predefined bias categories or any external data that can preserve essential semantic content without compromising overall accuracy.


To address these limitations, we propose \textbf{P}rojection-based \textbf{R}eduction of \textbf{I}mplicit \textbf{S}purious bias in vision-language \textbf{M}odels (PRISM). PRISM consists of two key stages. First, we prompt an LLM with labels to return a list of potential spurious connections corresponding to these labels and dynamically generate biased scene disruptions based on the extracted spurious attributes. Next, we find a projection of the embedding space that can effectively remove the impact of spurious attributes using the generated biased scene descriptions. 
To this end, we propose a novel loss named Latent space Debiasing loss ($\text{LD}$) to learn the debiasing projection in the embedding space effectively. Our proposed $\text{LD}$ loss is a contrastive-style debiasing loss that aligns text and image embeddings based on semantic meaning rather than spurious correlations. It ensures that the text embeddings of the same class but different spurious attributes remain similar to each other, and the text embeddings of different classes but the same spurious attribute remain distinct.

By removing the impact of bias attributes from the embedding space, PRISM generates representations with no or reduced statistical dependence on the bias. To measure this dependency, previously proposed methods mostly employ a metric capable of capturing statistical dependencies, such as Mutual Information \cite{reimers2021towards}, Hilbert-Schmidt Independence Criterion (HSIC) \cite{bahng2020learning}, or bias scores \cite{dehdashtian2024fairerclip}. In contrast, our proposed method eliminates the need for direct dependency measurement by utilizing an LLM to generate bias-aware descriptions. These descriptions capture common bias attributes in the dataset using only the labels.
Our approach facilitates debiasing without the need for predefined bias categories, external text or image datasets, or lossy data conversions to identify bias features. This makes PRISM an effective, task-agnostic, and data-free solution for debiasing VLMs.
 
We assess the performance of PRISM on CLIP using different benchmarks, including Waterbirds and CelebA datasets, and compare its performance against various baselines. The results highlight the capability of the proposed PRISM in mitigating spurious biases in VLMs without compromising the overall performance. 
Figure \ref{fig:short} illustrates that our proposed PRISM achieves state-of-the-art debiasing performance without compromising the total accuracy on both Waterbirds and CelebA benchmarks.  
Our key contributions in this paper can be summarized as follows:
\begin{itemize}
    \item We leverage LLMs to generate biased textual descriptions that reflect common spurious correlations found in datasets. This provides a dynamic and adaptable way to identify biases without relying on predefined bias categories.
    \item We propose a novel task-agnostic data-free debiasing method for VLMs that learns a projection of the embedding space, removing the impact of spurious attributes using a novel contrastive style loss function.  
    \item We demonstrate that our proposed method achieves state-of-the-art debiasing performance without compromising the model's overall accuracy. To allow rapid reproducibility and contribute to the community, we make our code public at: \url{https://github.com/MahdiyarMM/PRISM}.

\end{itemize}

\section{Related Works}
\noindent \textbf{Bias identification in VLMs.}
Recent research has placed considerable emphasis on identifying and assessing bias in foundation models  \cite{gallegos2024bias}, including VLMs \cite{agarwal2021evaluating, shao2023investigating}. The initial step in debiasing involves recognizing and evaluating the sources of bias. Different frameworks proposed in \cite{zeng2024understanding} and \cite{govil2024cobias, raza2024beads} comprehensively explored various types of biases in vision and text datasets, respectively. Previous works \cite{wang2022fairclip, agarwal2021evaluating, shao2023investigating} investigated bias in CLIP's representation space and proposed different methods to mitigate it. These biases cause unfair outcomes in zero-shot classification and reduce performance in certain underrepresented and overlooked categories \cite{shao2023investigating}.

\noindent \textbf{Debiasing VLMs.}
The methods proposed in \cite{zeng2024understanding, meister2023gender, alabdulmohsin2024clip} apply mathematical transformations on data to isolate bias in datasets before feeding them into models. However, such transformations may result in information loss. 
In the field of feature space transformation, FairerCLIP \cite{dehdashtian2024fairerclip} introduced a mathematically rigorous approach to reduce the bias in CLIP’s representations by leveraging kernel Hilbert spaces. SFID \cite{jung2025unified} proposed a selective feature imputation for debiasing, which combines feature pruning and low-confidence imputation to eliminate biases across tasks and modalities. These methods are grounded in rigorous mathematical principles for transformation but can be computationally expensive and risk removing crucial semantic information which compromises the overall accuracy of the model. 

In the field of prompt engineering, VisualDistiller \cite{dai2024refining} focused on refining visual representations to counter dataset biases. The solutions proposed in \cite{chuang2023debiasing}, namely Orth-cali and Orth-proj, efficiently remove the bias by projecting out the biased components from CLIP’s text embeddings using predefined biased prompts. However, because these biased prompts are fixed, the method may struggle to generalize to biases that weren't anticipated in the prompts. In the field of adapter designing, DFR \cite{kirichenko2022last} suggested debiasing by retraining only the model's final layer. On the other hand, contrastive adapter \cite{zhang2022contrastive} 
proposed to insert lightweight adapter modules into the pre-trained model architecture. These adapters are trained using a contrastive loss designed to pull representations from similar groups closer while pushing those from different groups apart. 

In a recent study, RoboShot \cite{adila2024zero} and BendVLM \citep{gerych2024bendvlm} also leveraged language models to extract insights from task descriptions. The primary focus of RoboShot was on improving robustness in underrepresented and adversarial data distributions rather than bias mitigation. BendVLM was proposed to tackle the bias in VLMs by learning task-specific projections via labeled images, with part of the projections computed dynamically during inference. In contrast, our PRISM explicitly targets bias mitigation in CLIP by leveraging LLMs for adaptive bias identification and embedding transformation.  PRISM eliminates the need for LLM calls during inference and removes spurious correlations through learning a debiasing projection of the embedding space, ensuring effective bias removal without relying on predefined bias categories or external visual data. Furthermore, unlike prompt-based debiasing techniques that modify only text embeddings, our approach preserves the alignment between text and image representations, ensuring debiased zero-shot classification without compromising the overall accuracy.

\section{Method}
\subsection{Problem formulation} \label{formulation}
Let us consider a multimodal foundation model, such as CLIP \cite{radford2021learning}, which maps images and natural language onto a shared $d$-dimensional space using two encoders: an image encoder $\phi_{\text{I}}(\cdot)$ and a text encoder $\phi_{\text{T}}(\cdot)$. In the zero-shot setting, given image $x$ and a set of text prompts $\{t_y\}_{y=1}^K$ corresponding to class labels $y \in \{1, \cdots, K\}$, the prediction is made via
\begin{equation}\label{zero}
\hat{y} = \arg\max_{y} \; \langle \phi_{\text{I}}(x), \phi_{\text{T}}(t_y) \rangle,
\end{equation}
where $\langle\cdot, \cdot\rangle$ represents the vector inner product. Here, the text prompts are descriptions of the object in each class as:  
$t_y=${\texttt{"A photo of a <class} $y$\texttt{>"}}.

\begin{figure*}[t]
  \centering
    \includegraphics[width=0.96\linewidth]{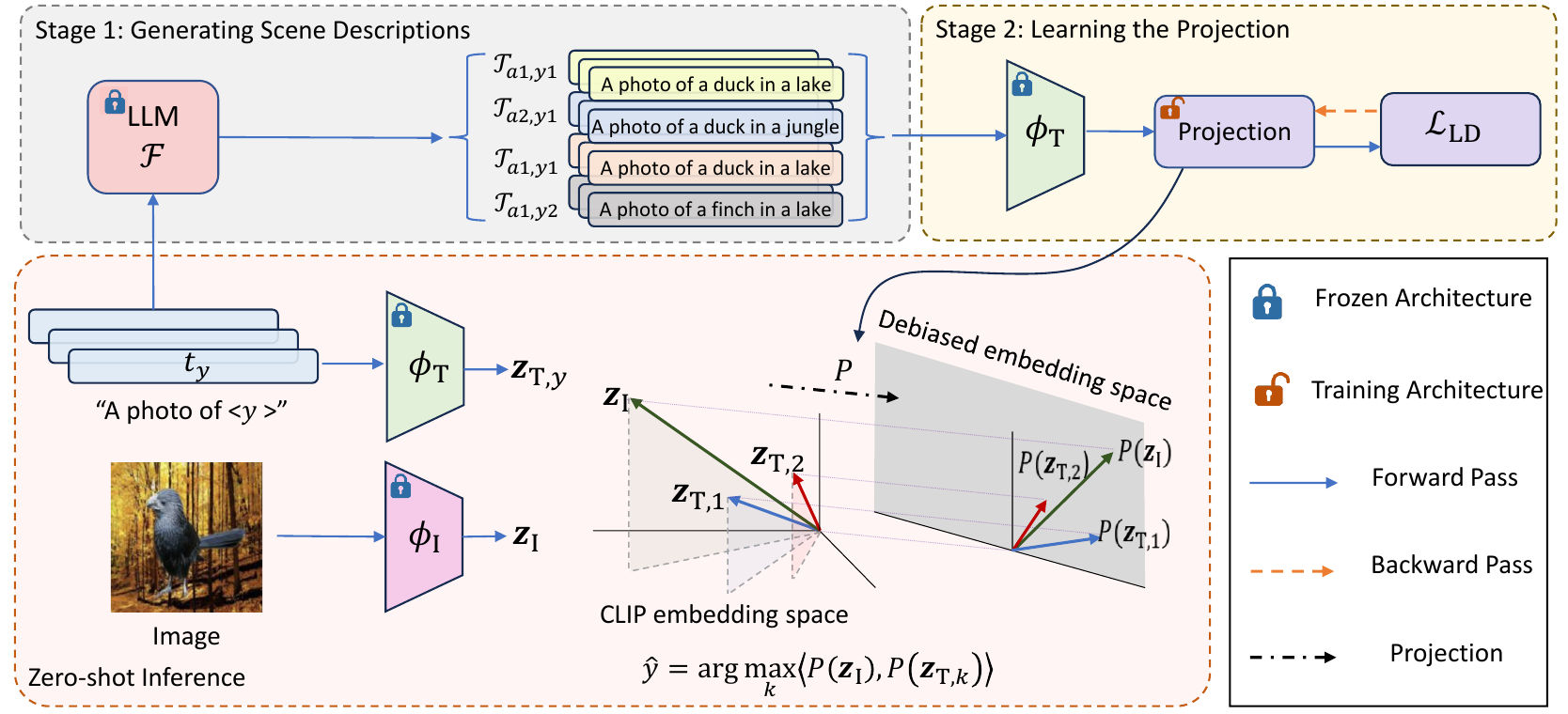}
  \caption{An overview of the proposed PRISM. In Stage 1, by passing text prompts as `A photo of $<y>$' to an LLM, a set of scene descriptions with different labels and spurious attributes is generated. Then, in Stage 2, these scene descriptions are used for learning a debiasing projection of the embedding space, $P$, using LD loss. Finally, zero-shot classification is performed in the projected debiased embedding space of the VLM.}
  \label{fig:clipper}
\end{figure*}

This formulation presumes that $\phi_I(x)$ predominantly captures the core, task-relevant features of $x$. However, real-world data often contain spurious correlations that can cause the encoder to also capture irrelevant cues. The image embedding can be modeled as a (normally, unknown) function $f$ of a core component and a spurious component:
\begin{equation}
\phi_{\text{I}}(x) = f(h_{\text{core}}(x), h_{\text{spu}}(x)),
\end{equation}
where $h_{\text{core}}(x)$ represents the features essential for correctly predicting the label $y$, and $h_{\text{spu}}(x)$ encodes extraneous information (e.g., background elements) that is spuriously correlated with $y$.

Assume that the spurious attribute is denoted by $a$. Accordingly, we define the \textit{group identity} of each sample as
\begin{equation}
g = (a, y), \quad \text{with} \quad g \in \mathcal{G} = A \times Y,
\end{equation}
where $A$ is the set of possible spurious attributes $a$, and $Y=\{1, \cdots, K\}$ corresponds to the set of true class labels. For example, in a bird classification task, $x$ is an image of a bird, and $y \in \{0, 1\}$ represents the true class where 0 denotes Landbird and 1 denotes Waterbird. The spurious attribute $a$ belongs to the set $A=\{{\texttt{land background}}, \, {\texttt{water background}}\}$, which indicates the type of background present in the image. Each combination $(a, y)$ defines a distinct group within $\mathcal{G}$, resulting in a total of four groups in this example. 

When the model undesirably relies on the spurious component $h_{\text{spu}}(x)$, it may assign $x$ to the incorrect group, leading to misclassification. To mitigate this bias, the objective is to minimize the worst-case error over all groups. Formally, we aim to solve
\begin{equation} \label{objective}
\min_{\theta \in \Theta} \; \max_{g \in \mathcal{G}} \; \mathbb{E}\Bigl[\ell\Bigl(\langle \phi_{\text{I}}(x), \phi_{\text{T}}(t_y)\rangle, y\Bigr) \,\Big|\, g\Bigr],
\end{equation}
where $\ell(\cdot, \cdot)$ denotes the classification loss. 

There are several challenges associated with the optimization of Eq. \ref{objective}. First, optimizing over the parameter space of $\theta$ is costly due to the number of parameters of the model, CLIP. This is also in contrast to the original purpose of CLIP which is to serve as a universal and general zero-shot classifier (Eq. \ref{zero}). Second, the identification of spurious attributes $A$ for an arbitrary set of text prompts is challenging and requires prior knowledge about the task, or an external dataset, which is not always accessible.
Moreover, we hypothesize that due to the nature of spurious correlations, similar biases should also be present in the text domain. Hence, spurious correlations identified in the text modality can also be leveraged to debias the image modality. We will empirically verify this hypothesis in Section \ref{sec:res}.

\subsection{PRISM}
To address the challenges mentioned in Section \ref{formulation}, 
we propose \textbf{P}rojection-based \textbf{R}eduction of \textbf{I}mplicit \textbf{S}purious bias in vision-language \textbf{M}odels (PRISM), which consists of two main steps: 
(\textbf{\textit{i}}) PRISM detects potential spurious correlations $A$ by prompting an LLM $\mathcal{F}$ with a set of inputs $\{t_y\}_{y=1}^K$ and instructs the model to generate a set of scene descriptions $\mathcal{T}$, corresponding to different groups $\mathcal{G}$.
(\textbf{\textit{ii}}) To preserve CLIP's generality and to avoid the need to fine-tune the network, we then introduce a novel method for finding an embedding projector $P$ that maps the embedding space onto another subspace where the influence of spurious attributes $A$ is minimized. Figure \ref{fig:clipper} presents a high-level illustration of our proposed method.

\noindent \textbf{Stage 1: Leveraging LLMs to identify spurious correlations.}
To propose a fully task-agnostic debiasing method without relying on any external data or predefined bias categories, we propose using pre-trained LLM for bias category discovery. For this purpose, an LLM is prompted with the labels and instructed to return potential spurious correlations. The LLMs' effectiveness in the discovery of spurious correlations stems from their training scheme. These foundation models are trained over a massive corpus $\mathcal{D}$ tokenized to $w_i$ by minimizing the following:
\begin{equation}
\frac{1}{N}\sum_{(w_1,\ldots,w_n)\in\mathcal{D}}\sum_{i=1}^n -\log P_\theta\bigl(w_i\,|\,w_{<i}\bigr), 
\end{equation}
thereby learning the joint distribution $P_\theta(w_1,\dots,w_n)$. Whenever a concept $\alpha$ (e.g., `camel') frequently co-occurs with another concept $\beta$ (e.g., `desert') in $\mathcal{D}$, the conditional probability $P_\theta(\beta \mid \alpha)$ becomes large, even if $\alpha$ and $\beta$ are merely spuriously correlated rather than causally connected. Hence, for any set of text prompts $\{t_y\}_{y=1}^K$, the LLM’s learned parameters encode high-probability transitions from $\{t_y\}_{y=1}^K$ to $A$ solely due to their co-occurrences in text \cite{sarkar2024data}. This statistical encoding of spurious correlations arises because the model optimizes the likelihood over $\mathcal{D}$ without discriminating between causal and non-causal patterns. Consequently, prompting the LLM with queries about $\{t_y\}_{y=1}^K$ produces $A$, as follows:
\begin{equation}
    A = \mathcal{F}(\{t_y\}_{k=1}^K).
\end{equation}
Then, we use the returned spurious attributes to generate a set of controlled scene descriptions $\mathcal{T}_{a,y}$ correspond to each group $g=(a,y)$ in $\mathcal{G}$. The full prompts used in this stage of PRISM are provided in Appendix B.

\noindent \textbf{Stage 2: Learning a debiasing projection of the embedding space.}
Our method finds a projection $P$ to minimize the impact of $A$ in the embedding space. This is then followed by zero-shot classification in the debiased projected embedding space.  
To this end, we propose a novel loss, which we call Latent space Debiasing Loss ($\text{LD}$), formulated as:
\begin{align}\label{loss}
&\mathcal{L}_{\text{LD}} 
= \frac{1}{K. \binom{|A|}{2}} 
  \underbrace{\sum_{\substack{(a,y)\in \mathcal{G}\\(a',y')\in \mathcal{G}\\a \neq a' \\y= y' }}
    \Bigl(
      1 - \langle \phi_{\text{T}}(\mathcal{T}_{a,y}), \phi_{\text{T}}(\mathcal{T}_{a',y'}) \rangle
    \Bigr)}_{\substack{\text{(a) Intra-class,} \\ \text{inter-attribute term}}}
    \nonumber 
+ \\ & \frac{1}{|A|. \binom{K}{2}}  \underbrace{\sum_{\substack{(a,y)\in \mathcal{G}\\(a',y')\in \mathcal{G}\\a = a' \\y\neq y'}}
    \max\Bigl\{
      0,\; \langle \phi_{\text{T}}(\mathcal{T}_{a,y}), \phi_{\text{T}}(\mathcal{T}_{a',y'}) \rangle - m
  \Bigr]}_{\substack{\text{(b) Inter-class,} \\ \text{intra-attribute term}}},
\end{align}
where $m$ is the margin hyperparameter. We propose $\mathcal{L}_{\text{LD}}$ to learn a projection that robustly mitigates spurious correlations in the model's joint embedding space. For each iteration of optimization over $P$,  we use a set of scene descriptions $\mathcal{T}$, generated by the LLM for each group, $g=(a,y)$.  
The first component of the loss ((a) Intra-class, inter-attribute term) encourages invariance by minimizing the difference between embeddings of scene descriptions from the same class but differing spurious attributes. Specifically, for each pair $(a,a')$ with $a \neq a'$, the term $1 - \langle \phi_{\text{T}}((a,y)), \phi_{\text{T}}((a',y')) \rangle$ is minimized. This term ensures that the core class-specific features are consistently represented, independent of the confounding spurious cues. Meanwhile, the second component of the loss ((b) Inter-class, intra-attribute term) employs a margin-based penalty that enforces a minimum separation between embeddings from the same spurious attribute but different classes. For any $y' \neq y$, the penalty $\max\{0,\, \langle \phi_{\text{T}}((a,y)), \phi_{\text{T}}((a',y')) \rangle - m\}$ is applied, ensuring that even if the spurious context is shared, class-discriminative features remain distinct. 

The design of $\mathcal{L}_{\text{LD}}$ is inspired by contrastive loss in self-supervised learning, where the objective is to draw semantically similar pairs closer and push dissimilar ones apart. By jointly optimizing for intra-class invariance and inter-class separation, our loss effectively penalizes deviations from the desired embedding structure. Importantly, this approach removes the need to fine-tune the entire CLIP model, as the learnable projection $P$ is optimized independently. Empirical evaluations confirm that minimizing $\mathcal{L}_{\text{LD}}$ not only enhances the robustness of the zero-shot classifier by reducing reliance on spurious cues but also improves worst-group performance and leads to disentanglement of the embedding space. 

Finally, leveraging the learned projection, $P$, by minimizing Eq. \ref{loss}, we perform zero-shot classification on the projected embedding space as follows:
\begin{equation}\label{zero_p}
\hat{y} = \arg\max_{k} \; \langle P(\phi_{\text{I}}(x)), P(\phi_{\text{T}}(t_k)) \rangle.
\end{equation}

\begin{algorithm}[t]
\caption{PRISM}
\small
\label{alg:debiasing}
\begin{algorithmic}[1]
\Require
\Statex image $x$
\Statex $\{t_y\}_{y=1}^K$: Text prompts for $K$ classes
\Statex $\phi_{\text{I}},\phi_{\text{T}}$: CLIP’s image and text encoders
\Statex $\mathcal{F}$: LLM
\vspace{1mm}
\State \textbf{Stage 1: Identify Spurious Directions}
\State $A = \mathcal{F}(\{t_y\}_{k=1}^K)$\quad $\triangleright$ Extract spurious attributes

\vspace{1mm}
\State \textbf{Stage 2 (PRISM): Learned Projection}
\State Generate scene descriptions $\mathcal{T}_{a,y}$ via $\mathcal{F}$ and $A$
\State Initialize $P$
\Repeat
   \State Compute $\mathcal{L}_{\text{LD}}$ (Eq. \ref{loss}) on $\{(a,y)\}$
   \State Update $P$ via gradient descent
\Until{converged}

\vspace{1mm}
\State \textbf{Stage 2 (PRISM-mini): Simple Orthogonal Projection}
\State $\mathcal{A} = \phi_\text{T}(A)$
\State $P \gets I - \mathcal{A} (\mathcal{A}^\top \mathcal{A})^{-1} \mathcal{A}^\top$

\vspace{1mm}
\State \textbf{Stage 3: Zero-shot Inference}
\State $\mathbf{z}_{\text{I}} \gets P(\phi_{\text{I}}(x)), \quad \mathbf{z}_{{\text{T}},k} \gets P(\phi_\text{T}(t_k))$
\State $\hat{y} \gets \arg\max_{k} \langle \mathbf{z}_{\text{I}}, \mathbf{z}_{{\text{T}},k}\rangle$
\end{algorithmic}
\end{algorithm}

\noindent \textbf{PRISM-mini.}
Here, we propose a variation of our method called PRISM-mini, which is mostly different from PRISM in stage 2. Unlike PRISM, which uses scene descriptions for finding $P$ by optimizing over LD loss, PRISM-mini relies on orthogonalizing the embedding space again spurious attributes. Hence, PRISM-mini is less computationally complex and is more suitable for use-cases where computation resources are extremely limited.

In the first stage, we pass the class prompts to the LLM and obtain a set of spurious attributes rather than full scene descriptions. Then, all these spurious attributes are encoded using CLIP's text encoder as follows:
\begin{equation} \label{LMM_detection}
\mathcal{A} = \phi_{\text{T}}(\mathcal{F}(\{t_k\}_{k=1}^K)),
\end{equation}
where $\mathcal{A}$ is a matrix whose columns are the embeddings of bias attributes in $A$. Inspired by \cite{chuang2023debiasing}, here we seek to reduce the impact of these spurious attributes on the embedding space by orthogonlizing it against $\mathcal{A}$. To this end, we calculate the projection $P$ as follows:
\begin{equation} \label{projection}
P = I - \mathcal{A}\,\bigl(\mathcal{A}^\top \mathcal{A}\bigr)^{-1}\,\mathcal{A}^\top\Bigr. .
\end{equation}
The rest of the method is essentially the same as PRISM. 
PRISM-mini can effectively reduce the impact of spurious correlations in CLIP without the need for any optimization for finding $P$. As a result, this method is very cost-effective. However, compared to PRISM, it might not be capable of fully mitigating the spurious biases in the embedding space due to its simpler projection mechanism. 
Algorithm \ref{alg:debiasing} summarizes both variations of the proposed method.

\begin{table*}[t]
\centering
\setlength{\tabcolsep}{7pt}
\resizebox{\linewidth}{!}{%
\begin{tabular}{l|ccccc|ccccc}
\toprule
\multirow{2}{*}{\textbf{Model}} 
& \multicolumn{5}{c|}{\textbf{Waterbirds}} 
& \multicolumn{5}{c}{\textbf{CelebA}} \\
& \textbf{WG} $\uparrow$ 
& \textbf{Acc} $\uparrow$ 
& $\Delta$\textbf{WG} $\uparrow$  
& $\Delta$\textbf{Acc} $\uparrow$ 
& \textbf{Gap} $\downarrow$
& \textbf{WG} $\uparrow$ 
& \textbf{Acc} $\uparrow$ 
& $\Delta$\textbf{WG} $\uparrow$ 
& $\Delta$\textbf{Acc} $\uparrow$
& \textbf{Gap} $\downarrow$ \\
\midrule \midrule
\rowcolor{lightblue} 
\multicolumn{11}{c}{\textbf{Baseline}} \\
\midrule
Zero-shot
& 36.4\% & 89.3\% & ---   & ---    & 52.9\% 
& 72.8\% & 87.6\% & ---   & ---    & 14.8\% \\
\rowcolor{lightblue} 
\midrule
\multicolumn{11}{c}{\textbf{Methods using images for debiasing}} \\
\midrule
WiSE-FT \cite{wortsman2022robust}
& 65.9\% & 97.6\% & 29.5\% & 8.3\%  & 31.7\%
& 80.0\% & 87.4\% & 7.2\%  & -0.2\% & 7.4\% \\
DFR (Sub) \cite{kirichenko2022last}
& 51.9\% & 95.7\% & 15.5\% & 6.4\%  & 43.8\%
& 76.3\% & 92.1\% & 3.5\%  & 4.5\%  & 15.8\% \\
DFR (Up) \cite{kirichenko2022last}
& 65.9\% & 96.1\% & 29.5\% & 6.8\%  & 30.2\%
& 83.7\% & 91.2\% & 10.9\% & 3.6\%  & 7.5\% \\
FairerCLIP \cite{dehdashtian2024fairerclip}
& 78.1\% & 85.1\% & 41.7\% & -4.2\% & \textbf{7.0\%}
& 86.1\% & 88.0\% & 13.3\% & 0.4\%  & \underline{1.9\%} \\
\rowcolor{lightblue} 
\midrule
\multicolumn{11}{c}{\textbf{Data-free methods}} \\
\midrule
VisualDistiller \cite{dai2024refining}
& 42.7\% & 90.6\% & 6.3\%  & 1.3\%  & 47.9\%
&  —    &  —    &  —     &  —     &  —     \\
Orth-Proj \cite{chuang2023debiasing}
& 45.3\% & 86.4\% & 8.9\%  & -2.9\% & 41.1\%
& 71.1\% & \textbf{87.0\%} & -1.7\% & \textbf{-0.6\%} & 15.9\% \\
Orth-Cali \cite{chuang2023debiasing}
& 68.8\% & 84.5\% & 32.4\% & -4.3\% & 15.7\%
& 76.1\% & 86.2\% & 3.3\%  & -1.4\% & 10.1\% \\
RoboShot \cite{adila2024zero}
& 45.2\% & 79.2\% & 8.8\%  & -10.1\%& 34.0\%
& \underline{82.6\%} & 85.5\%  & \underline{9.8\%}  & -2.1\% & 2.9\%  \\
PRISM-mini (ours)
& \underline{69.5\%} & \underline{92.6\%} & \underline{33.1\%} & \underline{3.3\%}  & 23.1\%
& \underline{82.6\%} & 84.4\%  & \underline{9.8\%}  & -3.2\% & \textbf{1.8\%} \\
PRISM (ours)
& \textbf{84.2\%} & \textbf{93.6\%} & \textbf{47.8\%} & \textbf{4.3\%}  & \underline{9.4\%}
& \textbf{84.0\%} & \underline{86.9\%} & \textbf{11.2\%} & \underline{-0.7\%} & 2.9\%  \\
\bottomrule
\end{tabular}%
}
\caption{The performance of PRISM compared to baselines on CLIP-ViT-L/14 on Waterbirds and CelebA datasets.  
The top performance in each column is in \textbf{bold}, and the second-best is \underline{underlined}.}
\label{tab:results_ViT}
\end{table*}

\section{Experiment setup}
\label{sec:exp}

\noindent \textbf{Datasets.} \label{secsec:data}
To evaluate the effectiveness of our method, 
we conduct various experiments on Waterbirds \cite{sagawa2019distributionally} and CelebA \cite{liu2015deep} datasets, following prior works such as \cite{dehdashtian2024fairerclip}. 
The Waterbirds dataset consists of images where bird species are paired with specific backgrounds, deliberately inducing spurious correlations between the foreground object and the environmental context. The CelebA dataset includes 200,000 facial images of celebrities annotated with 40 attributes, such as gender, hair color, face shape, etc.

\noindent \textbf{Evaluation.}
\label{secsec:Eval}
For performance evaluation, we use Five
metrics: Accuracy (Acc),  Worst-Group accuracy (WG), which is the lowest accuracy of all
subgroups, $\Delta$WG, which indicates the worst group accuracy improvement compared to the zero-shot clip, $\Delta$Acc, which indicates the overall accuracy improvement compared to the zero-shot performance of CLIP,  and the
Gap between Acc and WG.


\noindent \textbf {‌Baselines.} 
\label{secsec:Base}
We compare our method against two categories of baselines: the methods that use auxiliary images for debiasing and data-free methods. The first category employs a set of labeled or unlabeled images in the training process to reduce bias in CLIP. These images are used to keep the alignment of text and image embeddings intact throughout the debiasing process. Three state-of-the-art methods from this category are compared to ours, including WiSE-FT \cite{wortsman2022robust}, DFR \cite{kirichenko2022last}, and FairerCLIP \cite{dehdashtian2024fairerclip}. The second category consists of data-free models which do not require any images for debiasing. These methods are inherently task-agnostic as they do not rely on any task-specific visual data to achieve effective debiasing. VisualDistiller \cite{dai2024refining}, Orth \cite{chuang2023debiasing}, and RoboShot \cite{adila2024zero} are used in the following experiments as state-of-the-art methods in this category. Note that both variations of our method fall into this category.



\noindent \textbf{Implementation details.} \label{secsec:Imp}
All experiments were conducted using PyTorch on an NVIDIA GeForce RTX 3090 GPU with 24 GB VRAM and 64 GB RAM. Each experiment was run three times with different initialization seeds, and the average performance of the models was reported. Adam optimizer \cite{kingma2014adam} with the learning rate of 0.1 (Waterbirds) and 0.01 (CelebA) is used for performing optimization over $P$. 
The batch size is set to 64, and the training for the projection step is only performed for a single epoch.

\section{Results}\label{sec:res}

\noindent \textbf{Performance.}
We evaluate the performance of our method and compare it against prior baselines on two datasets (Waterbirds and CelebA) using the CLIP-ViT-L/14 backbone in Table \ref{tab:results_ViT}. The results demonstrate that PRISM outperforms prior works, setting a new state-of-the-art over data-free debiasing methods. Specifically, we observe that our proposed PRISM achieves the highest WG in both datasets, meaning it effectively reduces bias without needing additional image data. The effectiveness of our method is further demonstrated as PRISM outperforms the models using auxiliary data in the Waterbirds benchmark and achieve competitive results in the CelebA benchmark. 
FairerCLIP is the best among image-based methods, improving WG at the cost of some accuracy loss. WiSE-FT and DFR achieve high accuracies but have limited bias mitigation, indicating that fine-tuning might not be a good option to mitigate bias. Orth and RoboShot perform moderately well but struggle in WG gains, demonstrating that although prompt tuning is a viable method for guiding CLIP’s outputs, it is not a robust solution for bias removal. 
Moreover, the efficient variation of our method, PRISM-mini, also achieves the second-best performance in debiasing across all data-free methods while avoiding any optimization, making it a strong candidate for scenarios where resources are very limited.

\begin{figure*}[t]
  \centering
  \begin{subfigure}{0.49\linewidth}
    \includegraphics[width=\linewidth]{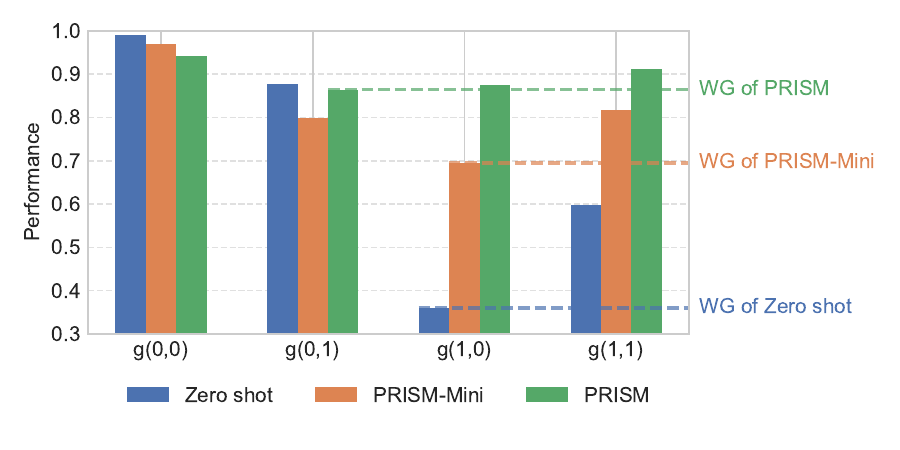}
    \caption{Waterbirds.}
    \label{fig:perG-a}
  \end{subfigure}
  \hfill
  \begin{subfigure}{0.49\linewidth}
    \includegraphics[width=\linewidth]{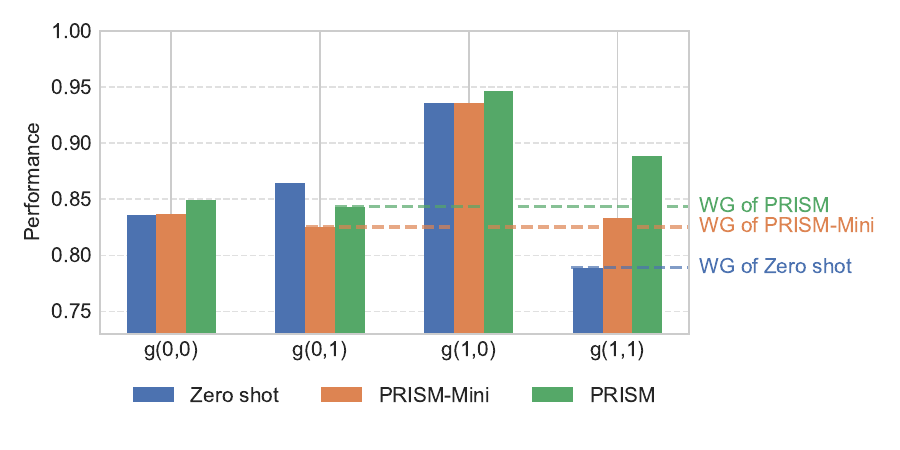}
    \caption{CelebA.}
    \label{fig:perG-b}
  \end{subfigure}
  \caption{Per-group accuracy 
  of our method on (a) Waterbirds and (b) CelebA datasets.}
  \label{fig:perG}
\end{figure*}


\begin{figure}
  \centering
  \begin{subfigure}{0.49\linewidth}
    \includegraphics[width=\linewidth]{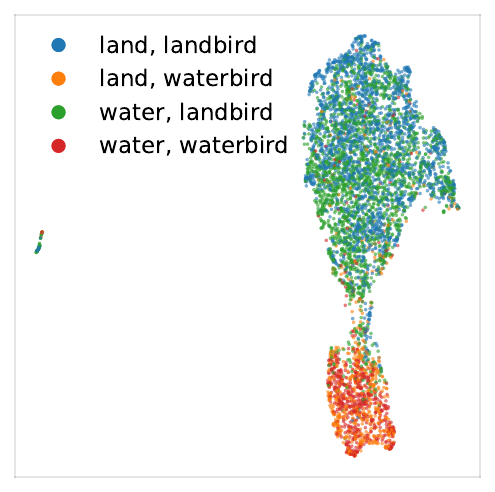}
    \caption{CLIP.}
    \label{fig:entengle-a}
  \end{subfigure}
  \hfill
  \begin{subfigure}{0.49\linewidth}
    \includegraphics[width=\linewidth]{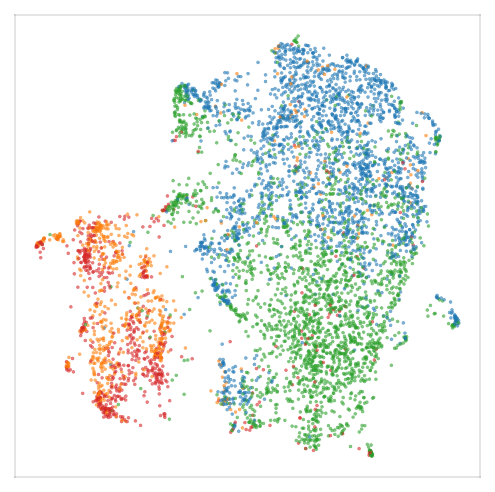}
    \caption{CLIP w/ PRISM.}
    \label{fig:entengle-b}
  \end{subfigure}
  \caption{The representation map at the output of CLIP (a) without and (b) with PRISM .}
  \label{fig:entengle}
\end{figure}

\noindent \textbf{Per-group accuracy.}
In Figure \ref{fig:perG}, each subfigure shows bar plots of per-group accuracy for three different methods, i.e., Zero-shot, PRISM-mini, and PRISM, on Waterbirds (Figure \ref{fig:perG-a}) and CelebA (Figure \ref{fig:perG-b}) datasets. The groups on the \textit{x}-axis ($g(0,0)$, $g(0,1)$, $g(1,0)$, $g(1,1)$) represent different subpopulations within each dataset, and the \textit{y}-axis shows the accuracy achieved by each method on those subpopulations. The dashed horizontal lines indicate the WG accuracy achieved by each method, i.e., the lowest accuracy among the four subgroups. By looking at where each bar stands relative to the dashed lines, we can see how each method handles the most challenging subgroup. Across both datasets, PRISM improves accuracy over the zero-shot baseline in each subgroup and tends to raise the WG performance. PRISM-mini also provides gains over the zero-shot baseline, but it may not always reach the same level of WG performance as PRISM. In essence, the figure illustrates that both PRISM variants can help reduce performance disparities across different groups compared to a purely zero-shot approach, with PRISM typically offering the strongest and the most balanced improvements.

\noindent \textbf{Bias text.}
We previously hypothesized that since CLIP aligns image and text embeddings, the inherent nature of spurious bias should cause similar biases to emerge in the text domain alone \cite{sarkar2024data}.
Consequently, spurious correlations extracted from the text domain can also be applicable to images. To empirically evaluate this hypothesis, we generate a set of scene descriptions featuring different bird species in various locations. We then perform zero-shot classification on the embeddings of these descriptions and present the results in Table \ref{tab:text_only}, demonstrating that similar biases can indeed be observed in the text domain.

\begin{table}[t]
\centering
\setlength{\tabcolsep}{3pt}
\resizebox{\columnwidth}{!}{%
\begin{tabular}{l|cc|cccc}
\toprule
{\textbf{Zero shot}} 
& \textbf{WG} $\uparrow$ 
& \textbf{Acc} $\uparrow$ 
& $g(0,0)$
& $g(0,1)$ 
&$g(1,0)$
&$g(1,1)$\\
\midrule
\midrule
Image
& 38.3\% & 92.8\%  & 99.0\% & 87.6\% & 38.3\% & 59.8\%
 \\
Scene description
& 33.1\% & 96.0\% & 100\% & 33.1\% & 41.9\% & 96.8\% \\
\bottomrule
\end{tabular}%
}
\caption{Bias similarity in text and image modalities.}
\label{tab:text_only}
\end{table}

\noindent \textbf{Trade-off analysis.}
Figures \ref{fig:tradeoff-a} and \ref{fig:tradeoff-b} compare different data-free CLIP debiasing methods in terms of accuracy and debiasing performance, measured through WG, on Waterbirds and CelebA datasets. Although some methods exhibit high debiasing performances, their overall performance is drastically degraded. PRISM is the most effective among all compared methods, achieving the best trade-off between these two metrics.  

\noindent \textbf{Feature disentanglement.} 
Figure \ref{fig:entengle} compares the representation distributions of zero-shot CLIP with PRISM on the Waterbirds dataset, where each color represents a group $(a,y) \in \mathcal{G}$. In CLIP's representation space (Figure \ref{fig:entengle-a}), clusters are highly mixed, indicating that the model heavily relies on spurious background correlations. In contrast, PRISM's representation space (Figure \ref{fig:entengle-b}) shows more structured and well-disentangled clusters, demonstrating that it effectively reduces the influence of background information. This analysis confirms that PRISM successfully mitigates bias rather than being influenced by spurious background correlations.

\begin{figure}
  \centering
  \begin{subfigure}{0.49\linewidth}
    \includegraphics[width=1.1\linewidth]{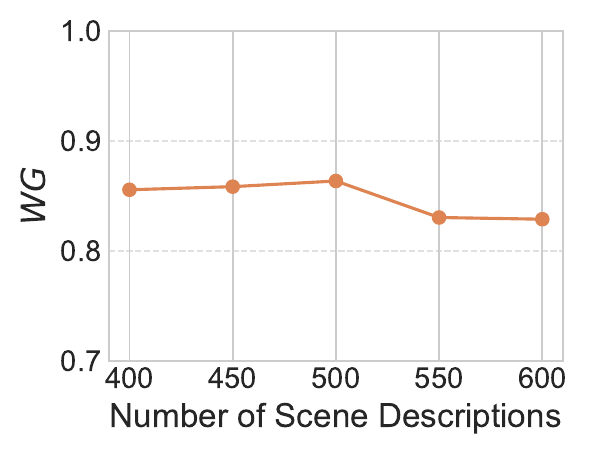}
    \caption{Waterbirds.}
    \label{fig:scene-a}
  \end{subfigure}
  \hfill
  \begin{subfigure}{0.49\linewidth}
    \includegraphics[width=1.1\linewidth]{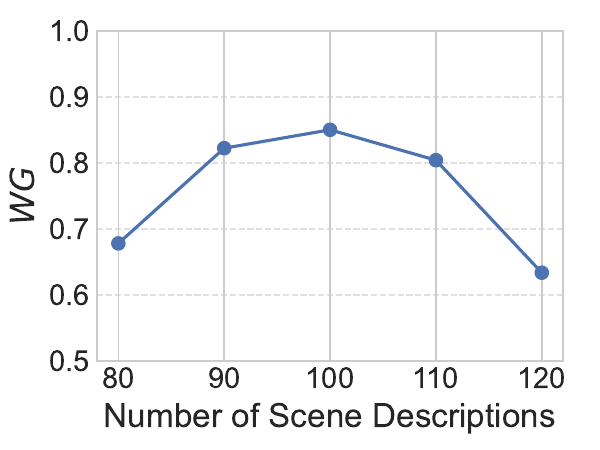}
    \caption{CelebA.}
    \label{fig:scene-b}
  \end{subfigure}
  \caption{Worst-group performance of PRISM for different numbers of scene descriptions used in training for (a) Waterbirds and (b) CelebA datasets.}
  \label{fig:scene}
\end{figure}

\noindent \textbf{Number of scene descriptions.}
Figure \ref{fig:scene} shows the effect of varying the number of scene descriptions used during training the projection $P$. In these experiments, we evaluate how increasing or decreasing the volume of scene descriptions, which serve as bias-aware prompts generated by the LLM, affects the model's ability to mitigate spurious correlations. For both the Waterbirds and CelebA datasets, the results illustrate that there is an optimal range for the number of descriptions: too few may provide insufficient guidance for debiasing, while too many might undermine the discriminative signal, ultimately impacting the worst-group accuracy. 

\begin{figure}
  \centering
  \begin{subfigure}{0.49\linewidth}
    \includegraphics[width=1.1\linewidth]{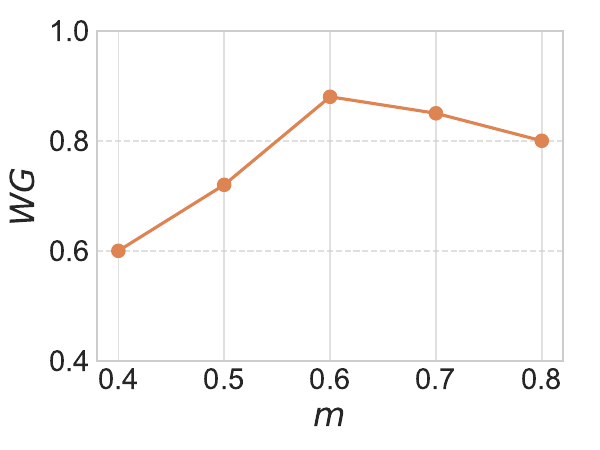}
    \caption{Waterbirds.}
    \label{fig:m-a}
  \end{subfigure}
  \hfill
  \begin{subfigure}{0.49\linewidth}
    \includegraphics[width=1.1\linewidth]{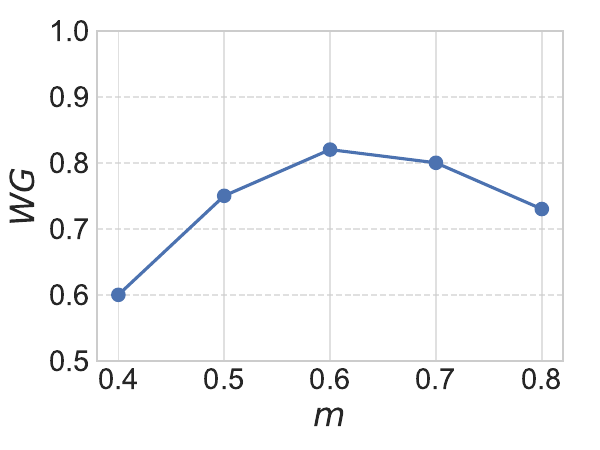}
    \caption{CelebA.}
    \label{fig:m-b}
  \end{subfigure}
  \caption{Worst-group performance of PRISM for different values of margin hyperparameter $m$ for (a) Waterbirds and (b) CelebA datasets.}
  \label{fig:m}
\end{figure}

\noindent \textbf{$\bm{m}$ in $\text{LD}$ loss.}
Figure \ref{fig:m} shows the sensitivity of PRISM’s performance to the hyperparameter $m$ 
in the $\text{LD}$ loss function, 
which determines the margin in the contrastive component of the loss. This parameter influences how aggressively the model penalizes similarities between embeddings of different classes with the same spurious attribute. The results, 
demonstrate 
that for both datasets, an $m$ of 0.6 maximizes WG performance 
by separating the representations across different groups, achieving an optimal balance between preserving essential semantic information and removing spurious correlations.

\noindent \textbf{Role of LLMs.}
Table \ref{tab:llm} shows that the effectiveness of PRISM in reducing bias, measured by WG and Acc, can be significantly influenced by the choice of LLM. Selecting a more capable LLM for bias identification can improve debiasing without severely compromising the overall classification performance. The table compares several LLMs that vary in size and capability, where we see that GPT‑4o achieves the best WG performance with score of $84.0\%$ and an overall accuracy of $86.9\%$.

\begin{table}[t]
\vspace{4mm}
\centering
\setlength{\tabcolsep}{5pt}
\resizebox{0.75\columnwidth}{!}{%
\begin{tabular}{l|cc}
\toprule
{\textbf{LLM}} 
& \textbf{WG} $\uparrow$ 
& \textbf{Acc} $\uparrow$ \\
\midrule
\midrule
Llama 3.2 (70B) \cite{dubey2024llama}& 58.9\%& 90.1\% \\
DeepSeek r1 (8B) \cite{guo2025deepseek} & 68.3\%&89.2\% \\
GPT 4o mini ($\sim$8B) \cite{hurst2024gpt} & 81.6\%&86.6\% \\
GPT o1 ($\sim$300B) \cite{jaech2024openai} &77.9\% &83.9\% \\
GPT 4o ($\sim$1.7T) \cite{hurst2024gpt} &84.0\% & 86.9\% \\
\bottomrule
\end{tabular}%
}
\caption{The performance of PRISM on CelebA dataset using different LLMs.}
\label{tab:llm}
\end{table}

\noindent \textbf{Limitations.}
While PRISM demonstrates strong performance in mitigating spurious biases in VLMs, it has some limitations. First, PRISM relies on LLMs to generate spurious attribute descriptions, and the quality of these descriptions can vary depending on the specific LLM used. While our results show that more capable LLMs yield better debiasing performance, less powerful models may struggle to extract meaningful spurious correlations, potentially reducing effectiveness. Additionally, our approach focuses on linear projections to remove bias in the embedding space, which may not always capture highly complex or non-linear biases present in real-world data.

\section{Conclusion}
We introduced PRISM, a novel data-free debiasing framework for VLMs that effectively mitigates spurious correlations in CLIP’s embedding space. PRISM has two stages. In the first stage, an LLM is leveraged with simple class prompts
to generate scene descriptions that 
unravel
common spurious correlations in the training data. In the second stage, these generated descriptions guide the learning of an optimized projection mapping, which is carried out via our novel LD loss to systematically reduce spurious biases while preserving the semantic alignment between image and text embeddings. 
Through extensive evaluations on benchmark datasets (Waterbirds and CelebA), we demonstrated that PRISM significantly improves worst-group accuracy while maintaining overall classification performance. Beyond its empirical effectiveness, PRISM offers a 
data-free and task-agnostic solution to debiasing multimodal representations. 
Future research directions include exploring non-linear projections for embedding space debiasing, which could potentially better capture complex spurious correlations beyond linear transformations.

\section*{Acknowledgment}
We would like to thank Geotab Inc., the City of Kingston, and NSERC for their invaluable and continued support of this work.

\bibliographystyle{ieeenat_fullname}
\bibliography{main}

\clearpage    

\onecolumn    
\appendix

\renewcommand{\thetable}{A\arabic{table}}
\setcounter{table}{0}

\begin{center}
  {\huge\bfseries Appendix}
\end{center}
\vspace{1em}

\section{Notation}\label{app:notation}
Here, we summarize the key notations used throughout the paper:
\begin{itemize}
    \item $x$: An input image.
    \item $\{t_y\}_{y=1}^K$: A set of $K$ text prompts corresponding to class labels.
    \item $\phi_{\text{I}}(\cdot)$: CLIP’s image encoder that maps an image to a $d$-dimensional embedding.
    \item $\phi_{\text{T}}(\cdot)$: CLIP’s text encoder that maps text to a $d$-dimensional embedding.
    \item $d$: Dimensionality of the shared embedding space.
    \item $\langle \cdot, \cdot \rangle$: Inner product in the embedding space.
    \item $\hat{y}$: Predicted class label computed as 
    \[
    \hat{y} = \arg\max_{k}\; \langle \phi_{\text{I}}(x), \phi_{\text{T}}(t_k) \rangle.
    \]
    \item $h_{\text{core}}(x)$: The core features of image $x$ that are essential for correct classification.
    \item $h_{\text{spu}}(x)$: The spurious features of image $x$ that may capture irrelevant cues.
    \item $f$: A function that combines core and spurious features, i.e., 
    \[
    \phi_{\text{I}}(x) = f\bigl(h_{\text{core}}(x), h_{\text{spu}}(x)\bigr).
    \]
    \item $a$: A spurious attribute (e.g., background type).
    \item $y$: The true class label.
    \item $g = (a, y)$: The group identity, with 
    \[
    g \in \mathcal{G} = A \times Y,
    \]
    where $A$ is the set of spurious attributes and $Y=\{1,\dots,K\}$ is the set of class labels.
    \item $\mathcal{T}_{a,y}$: Scene descriptions corresponding to $(a,y)$
    \item $\ell(\cdot,\cdot)$: The classification loss function.
    \item $\theta \in \Theta$: The parameters of the model.
    \item $\mathcal{F}$: A pre-trained large language model (LLM) used to detect spurious attributes.
    \item $\mathcal{A}$: The embedding of spurious attributes, computed as 
    \[
    \mathcal{A} = \phi_{\text{T}}(\mathcal{F}(\{t_y\}_{k=1}^K)).
    \]
    \item $P$: A projection operator applied to the embedding space to reduce the impact of spurious cues.
    \item $\mathcal{L}_{\text{LD}}$: The Latent space Debiasing Loss used to learn $P_{\text{learn}}$, which balances intra-class invariance and inter-class separation.
    \item $m$: The margin hyperparameter in $\mathcal{L}_{\text{LD}}$.
\end{itemize}

\section{Prompts}
To find the spurious correlations, we use the following prompt:
\texttt{"Provide a list of potential bias attibutes associated with the following zero-shot classification using CLIP:
<$\{t_k\}$>"}.

\noindent Then, we use generate a set of controlled scene descriptions using the following:
\texttt{"based on these classes <$\{t_k\}$> and these spurious attributes
<$A$>, create <$n$> scene descriptions where the class and attributes are marked as "*class*" and "*attribute*" and can be later replaced with a class or attribute from its corresponding list. Generate a list for each class and a list for each attributes separately.
Use this example as a guide:
<"Example of\\Panda and Camel with spurious connection with Desert and Bamboo">"}.

\begin{table*}[t]
\centering
\setlength{\tabcolsep}{7pt}
\resizebox{\linewidth}{!}{%
\begin{tabular}{l|ccccc|ccccc}
\toprule
\multirow{2}{*}{\textbf{Model}} 
& \multicolumn{5}{c|}{\textbf{Waterbirds}} 
& \multicolumn{5}{c}{\textbf{CelebA}} \\
& \textbf{WG} $\uparrow$ 
& \textbf{Acc} $\uparrow$ 
& $\Delta$\textbf{WG} $\uparrow$  
& $\Delta$\textbf{Acc} $\uparrow$ 
& \textbf{Gap} $\downarrow$
& \textbf{WG} $\uparrow$ 
& \textbf{Acc} $\uparrow$ 
& $\Delta$\textbf{WG} $\uparrow$ 
& $\Delta$\textbf{Acc} $\uparrow$ 
& \textbf{Gap} $\downarrow$ \\
\midrule \midrule
\rowcolor{lightblue} 
\multicolumn{11}{c}{\textbf{Baseline}} \\
\midrule
Zero-shot
& 38.3\% & 92.8\% & –     & –      & 54.5\%
& 75.8\% & 82.4\% & –     & –      & 6.6\% \\
\rowcolor{lightblue} 
\midrule
\multicolumn{11}{c}{\textbf{Methods using images for debiasing}} \\
\midrule
DFR (Sub) [{\color{iccvblue}16}]
& 66.1\% & 92.9\% & 27.8\% & 0.1\%  & 26.8\%
& 80.9\% & 91.7\% & 5.1\%  & 9.3\%  & 10.8\% \\
DFR [{\color{iccvblue}16}]
& 54.2\% & 90.3\% & 15.9\% & –2.5\% & 36.1\%
& 89.9\% & 91.3\% & 14.1\% & 8.9\%  & \textbf{1.4\%} \\
FairerCLIP [{\color{iccvblue}7}]
& 75.4\% & 84.3\% & 37.1\% & –8.5\% & \underline{8.9\%}
& 81.5\% & 85.0\% & 5.7\%  & 2.6\%  & 3.5\% \\
\rowcolor{lightblue} 
\midrule
\multicolumn{11}{c}{\textbf{Data-free methods}} \\
\midrule
VisualDistiller [{\color{iccvblue}6}]
& 44.2\% & \textbf{93.1}\% & 5.9\%  & \textbf{0.3}\%  & 48.9\%
& –      & –      & –     & –      & –     \\
Orth-Proj [{\color{iccvblue}5}]
& 48.1\% & 83.6\% & 9.8\%  & –9.2\% & 35.5\%
& 61.4\% & \underline{86.4}\% & –14.4\% & \underline{4.0\%}  & 25.0\% \\
Orth-Cali [{\color{iccvblue}5}]
& \textbf{74.0}\% & 78.7\% & \textbf{35.7\%} & –14.1\% & \textbf{4.7\%}
& \underline{82.2}\% & 84.4\% & \underline{6.4\%}  & 2.0\%  & \underline{2.2\%} \\
PRISM-Mini (ours)
& 62.0\% & 91.7\% & 23.7\% & –1.1\% & 29.7\%
& 70.3\% & 79.2\% & –5.5\% & –3.2\% & 8.9\% \\
PRISM (ours)
& \underline{70.2}\% & \underline{91.9}\% & \underline{31.9\%} & \underline{–0.9\%} & 21.7\%
& \textbf{83.3}\% & \textbf{89.0\%} & \textbf{7.5\%}  & \textbf{6.6\%}  & 5.7\% \\
\bottomrule
\end{tabular}%
}
\caption{The performance of PRISM compared with baselines on CLIP-RN50 on Waterbirds and CelebA datasets.  
Top performers in each column are \textbf{bolded}, second bests are \underline{underlined}.}
\label{tab:results_RN}
\end{table*}

\section{Additional experiments}

Table \ref{tab:results_RN} presents the results for the CLIP-RN50 model, including only the baselines that report for CLIP-RN50. On the CelebA dataset, PRISM attains the best performance in both WG and Acc metrics. However, it does not outperform the state-of-the-art methods on the Waterbirds dataset. This is partly because Waterbirds is smaller in scale and has a stronger background–label correlation, making a less powerful backbone like ResNet-50 more susceptible to entangling these features. In this case, finding spurious features without any predefined prompts, is more difficult for PRISM.

\begin{table*}[t]
\centering
\setlength{\tabcolsep}{7pt}
\resizebox{\linewidth}{!}{%
\begin{tabular}{l|ccccc|c}
\toprule
\multirow{2}{*}{\textbf{Model}} 
& \textbf{Attribute Labels} 
& \textbf{Task Info at Inf.} 
& \textbf{LLM at Inf.} 
& \textbf{Inference Speed} 
& \textbf{Explicit Debiasing} 
& \textbf{Worst Group AUROC} \\
& (Need) & (Need) & (Need) & (Relative) & (Objective) & (CelebA \%) \\
\midrule \midrule
\rowcolor{lightblue} 
\multicolumn{7}{c}{\textbf{Comparison of PRISM variants and BendVLM}} \\
\midrule
Zero-shot
& No & No & No & Fast & None & 72.8 \\
BendVLM [{\color{iccvblue}10}]
& Yes & Yes & Yes & Slow & Yes (inference constraint) & 77.2 \\
PRISM-mini (ours)
& No & No & No & Fast & Yes (simple projection) & 82.6 \\
PRISM (ours)
& No & No & No & Fast & Yes (LD loss) & \textbf{84.0} \\
\bottomrule
\end{tabular}%
}
\caption{Comparison of Zero-shot CLIP, PRISM-mini, PRISM, and BendVLM in terms of practical requirements, design differences, and performance on the CelebA dataset.}
\label{tab:prism_vs_bendvlm}
\end{table*}

To further highlight the effectivenss of our method, We also compare PRISM with BendVLM. While both PRISM and BendVLM aim to mitigate biases in vision-language models, they differ fundamentally in several aspects that make PRISM more practical, efficient, and effective. Specifically: (i) PRISM does not require any labeled image data for training, unlike BendVLM which relies on annotated images. (ii) PRISM needs no task-specific information (e.g., class labels or number of classes) at inference time. (iii) PRISM is faster at inference by applying a single fixed projection learned during training, whereas BendVLM recomputes a projection per test input. (iv) PRISM requires LLM access only once during training, while BendVLM needs repeated LLM queries at inference, resulting in higher cost and latency. (v) PRISM explicitly introduces an LD loss that enforces intra-class invariance and inter-class separability. (vi) As summarized in Table X, PRISM achieves better performance across key benchmarks.

\end{document}